\documentclass{article}

\usepackage{ijcai18}

\usepackage{times}
\usepackage{xcolor}
\usepackage{soul}
\usepackage[utf8]{inputenc}
\usepackage[small]{caption}

\usepackage{graphicx}
\usepackage{amsmath}
\usepackage{amssymb}
\usepackage{siunitx}
\usepackage{textcomp}
\usepackage{color}

\usepackage{makecell}

\title{Layered Optical Flow Estimation Using a Deep Neural Network with a Soft Mask}

\author{
Xi Zhang, 
Di Ma, 
Xu Ouyang,
Shanshan Jiang,
Lin Gan,
Gady Agam, 
\\ 
Computer Science Department, Illinois Institute of Technology \\
\{xzhang22, dma2, xouyang3, sjiang20\}@hawk.iit.edu,
\{lgan, agam\}@iit.edu
}

\begin{document}

\maketitle

\begin{abstract}
Using a layered representation for motion estimation has the advantage of being able to cope with discontinuities and occlusions. In this paper, we learn to estimate optical flow by combining a layered motion representation with deep learning. Instead of pre-segmenting the image to layers, the proposed approach automatically generates a layered representation of optical flow using the proposed soft-mask module. The essential components of the soft-mask module are maxout and fuse operations, which enable a disjoint layered representation of optical flow and more accurate flow estimation. We show that by using masks the motion estimate results in a quadratic function of input features in the output layer. The proposed soft-mask module can be added to any existing optical flow estimation networks by replacing their flow output layer. In this work, we use FlowNet as the base network to which we add the soft-mask module. The resulting network is tested on three well-known benchmarks with both supervised and unsupervised flow estimation tasks. Evaluation results show that the proposed network achieve better results compared with the original FlowNet. 

\end{abstract}

\section{Introduction}

Optical flow estimation is a crucial and challenging problem with numerous applications in computer vision. Traditional differential methods for estimating optical flow include variational methods~\cite{horn1981determining}, which uses a regularization term and provide a global solution for optical flow. Various improvements of these initial formulations have been proposed over many years.

\begin{figure}[h]
\centerline{
\begin{tabular}{c}
  \resizebox{0.4\textwidth}{!}{\rotatebox{0}{
  \includegraphics{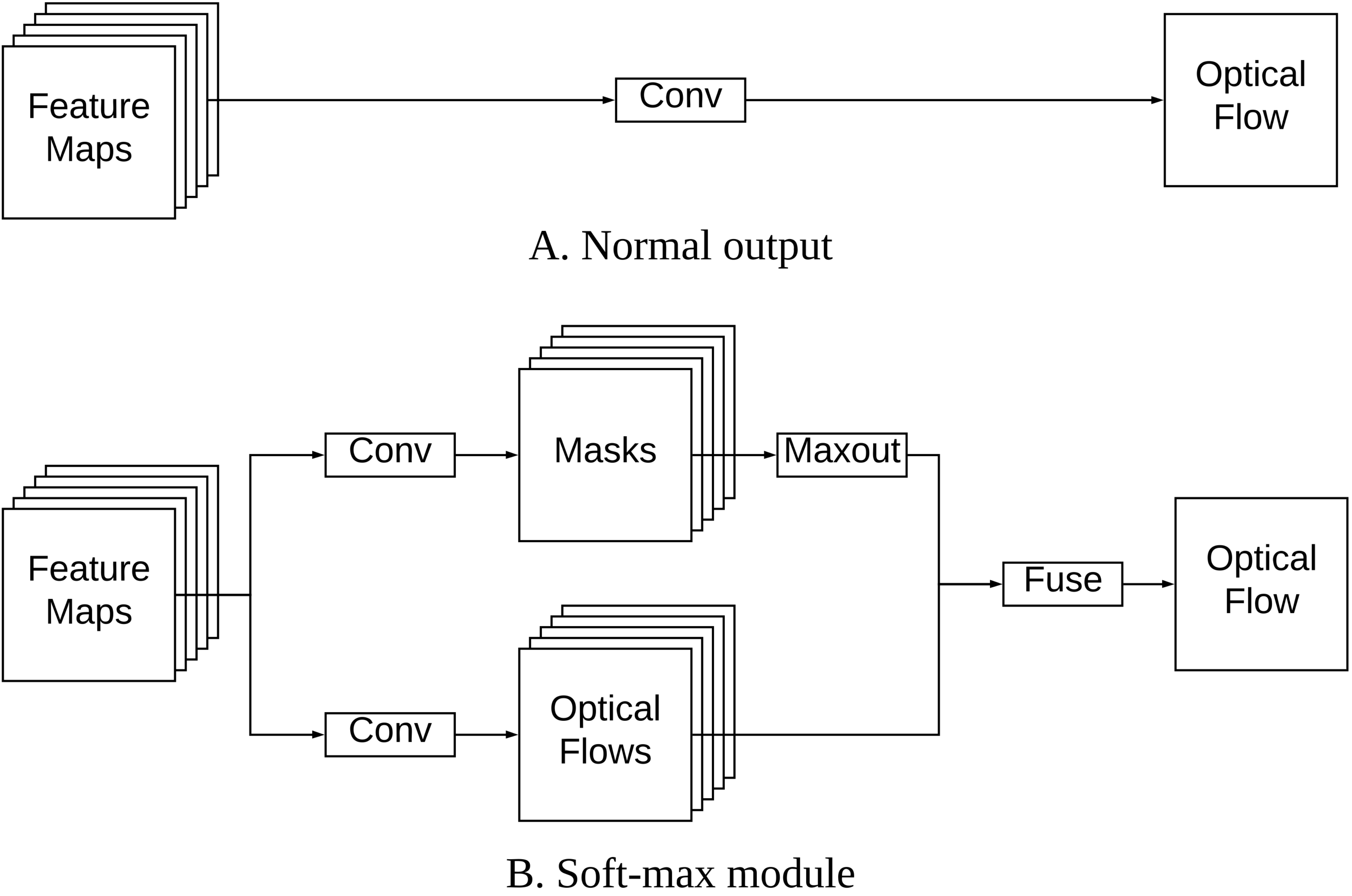}}}
\end{tabular}}
\caption{Illustration of the structure of the proposed soft-mask module compared with traditional linear optical flow network.}
\label{fig: soft-mask module}
\end{figure} 

Layered models of optical flow offer an easy performance boost for optical flow estimation. Disjointly splitting the optical flow into layers enables easier modeling of optical flow in each layer. Such a representation is especially helpful for small object motion estimation, as many optical flow estimation techniques are biased towards motion in large areas. Layered representation also improves flow computation on flow field boundaries by handling the smoothness constraint separately in each layer.

FlowNet~\cite{7410673} was the first to use a deep neural network for end-to-end optical flow estimation. FlowNet is fundamentally different from established differential approaches. As traditional differential optical flow estimation techniques perform well and are well established, several deep learning based approaches try to bridge the gap between traditional approaches and deep learning based approaches by using the best of both sides. For example, Ranjan et al.~\cite{Ranjan_2017_CVPR} use a pyramid representation of flow and residual flows to address large flow displacement estimation. Several approaches~\cite{ren2017unsupervised}\cite{ahmadi2016unsupervised}\cite{DBLP:journals/corr/YuHD16} investigated the basic principles of flow estimation and proposed unsupervised network training. 

\begin{figure*}[th]
\centerline{
\begin{tabular}{c}
  \resizebox{0.51\textwidth}{!}{\rotatebox{0}{
  \includegraphics{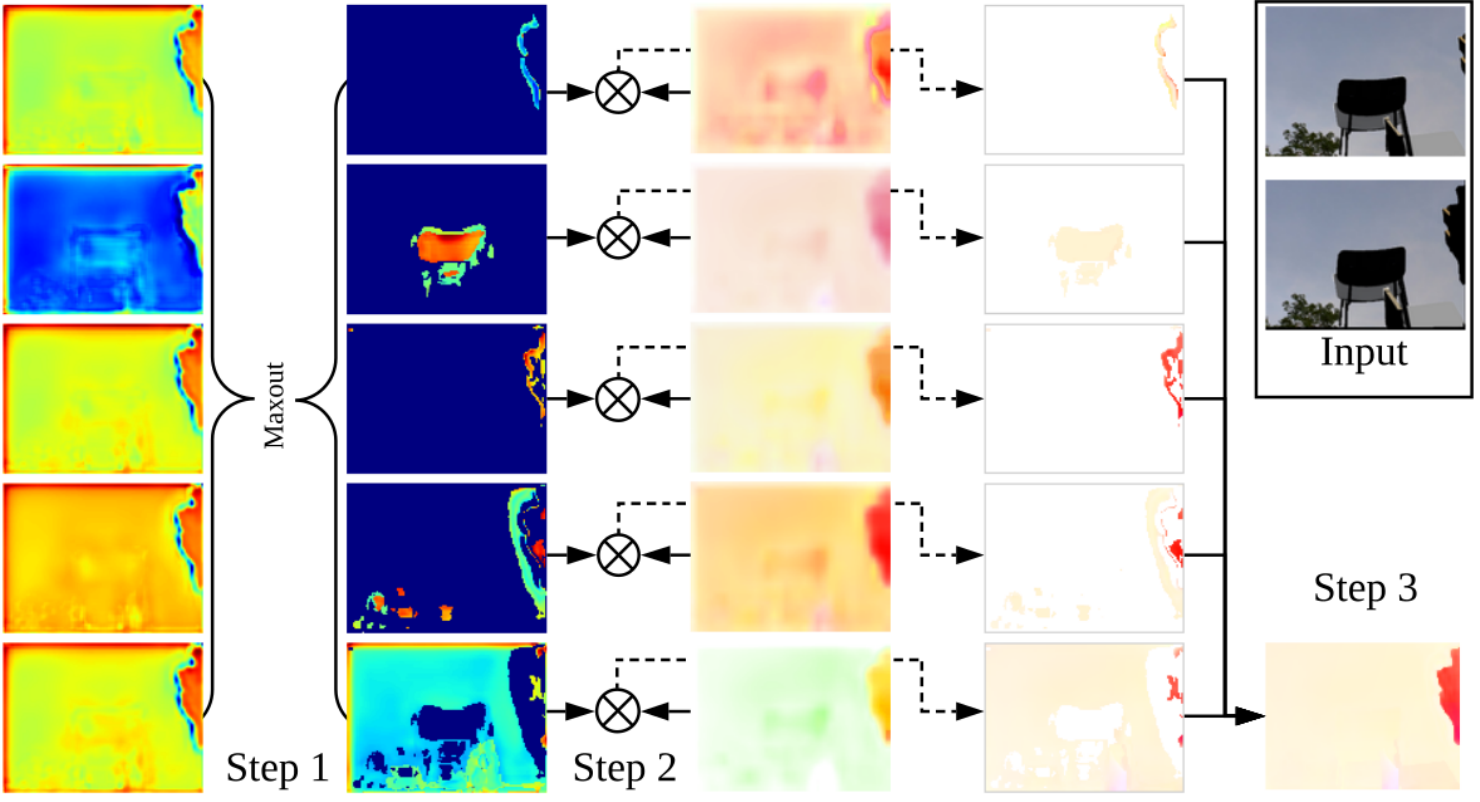}}}
  \\
\end{tabular}}
\caption{Pipeline of the soft-mask module. Step 1: Setting maxout and non-max pixels in masks to zero. Step 2: Pixel-wise multiplication with generated intermediate flows. Step 3: Generating final flow by summing up flows from every mask.}
\label{fig: softmask pipeline}
\end{figure*} 

Our work  \footnote[1]{This is paper is accepted by IJCAI 2018} combines the idea of using a layered optical flow representation with a deep neural network structure. Unlike previous approaches~\cite{yang2015dense}, where the layered representation is generated separately, the layered representation in the proposed approach is inferred internally and automatically when training the neural network. We achieve this by designing a soft-mask module. The soft-mask module is a network structure which splits optical flow into layers using disjoint real-valued masks. As the masks are not binary, we use the term 'soft' to refer to them. The soft-mask module offers a more accurate flow estimation due to two unique characteristics. The first is its ability to represent estimated flow using disjoint layers, which results in a more focused and simpler flow estimation for each layer. Second, compared with the linear flow output in FlowNet, the flow estimated using the soft-mask module is quadratic in terms of input features. This allows the soft-mask module to better fit more complicated optical flow patterns. The idea of using the soft-mask module is similar to the maxout networks proposed by Goodfellow~\cite{Goodfellow:2013:MN:3042817.3043084}, where the output of a neuron is the max of a set of inputs. The proposed soft-mask module extends the maxout operation to 2D.  In addition, instead of keeping max values only, we zero-out non-max values and use them when fusing layered optical flows.

In this work, the soft-mask module is added to FlowNet by replacing the output layer of the network with the soft-mask module. While more generally, the soft-mask module could be used in other per-pixel prediction tasks such as semantic segmentation and single image depth estimation, we focus in this paper on its application to optical flow estimation.

We show that by using the soft-mask module, we boost the performance of FlowNet when tested on several public datasets such as the Flying Chairs~\cite{7410673},  Sintel~\cite{Butler:ECCV:2012}, and KITTI~\cite{geiger2012we}.  We further show that both supervised and unsupervised flow estimation methods benefit from using the soft-mask module.

\subsection{Related Work}
Our work effectively combines ideas from using layered representation in classical optical flow approaches with recent deep learning approaches.
\newline
\newline
\noindent \textbf{Layered approaches.}
Using layered approaches in motion estimation are commonly used to overcome discontinuities and occlusions. A layered approach has been proposed~\cite{darrell1991robust} where a Bayesian model for segmentation and robust statistics is incorporated.  Recent work~\cite{sun2010layered} use affine motion to regularize the flow in each layer, while Jepson and Black~\cite{341161} formalize the problem using probabilistic mixture models. Yang~\cite{yang2015dense} fit a piecewise adaptive flow field using piecewise parametric models while maintaining a global inter-piece flow continuity constraint. Exploiting recent advances in semantic scene segmentation, \cite{Sevilla-Lara_2016_CVPR} use different flow types for segmented objects in different layers. Hur and Roth~\cite{Hur2016} treat semantic segmentation and flow estimation as a joint problem.
\newline
\newline
\noindent \textbf{Deep learning approaches.}
Deep neural networks have been shown to be successful in many computer vision tasks including object recognition and dense prediction problems~\cite{long2015fully}. FlowNet attempts to solve optical flow estimation using a deep neural network. FlowNet provides an end-to-end optical flow learning framework which serves as a base model for many later works. Two notable existing works include \cite{zhou2016view} and \cite{flynn2016deepstereo}. Masks generated by these approaches are normalized and then used as weight maps and multiplied with features before a final prediction is made. The masks used in our proposed work is different from masks in prior works. Instead of normalizing mask values across different channels, the proposed soft-mask module applies a maxout operation among channels. Because of the maxout operation, the soft-mask module could segment objects better with different flows. In  Section~\ref{sec: evaluation of the soft-mask module}, we compare the results of different ways of using masks in optical flow prediction and verify that the proposed soft-mask module over performs other existing methods.

\subsection{Novel Contribution}
In this work, we extend FlowNet and improve its performance in several ways. First, we propose combining a traditional layered approach for optical flow estimation with deep learning. The proposed approach does not require pre-segmentation of images. Instead, the separation of layers is done automatically when training the network. Second, a soft-mask module is proposed. This soft-mask module implements a channel-wise maxout operation among masks. As a result, the estimated optical flow is separated into layers, each of which contains optical flow that is estimated using a quadratic function. Third, we extend FlowNet by adding the proposed soft-mask module in the output layers. The resulting network is trained and compared with both supervised and unsupervised optical flow estimation approaches using neural networks. Experimental results show that the proposed network structure achieves lower error in each experimental group.

\section{Methodology}
%
%

\subsection{Soft-mask Module}
FlowNet was the first work to use a deep convolutional neural network for optical flow estimation. The network architecture used by FlowNet is very similar to the structure of a classical auto-encoder, where optical flows are generated using deconvolution at each scale level of the image pyramid. To refine flow estimations, shortcuts are built to connect layers of corresponding levels in the encoder and decoder layers. Consider a single computation of convolution, and for simplicity assume that $f$ represents both horizontal and vertical components of the output flow. Given $X \in \mathbb{R}^{s\times s \times c}$, representing an input feature volume, where $s$ is the kernel size, and $c$ is the number of channels, FlowNet employs a linear activation to compute optical flow:

\begin{equation}
f = X^T W + b
\end{equation}


Given that actual optical flow fields are nonlinear and piecewise smooth, using a linear function to fit the flow field shifts the non-linearity to the convolutional layers making the learning there more difficult. Using the soft-mask module proposed in this paper to replace the linear output of optical flow estimation, we can separate the optical flow field into multiple layers. The flow estimation in each layer is smooth and is easier to estimate compared with the original model. This results in a more accurate and flexible optical flow estimation. 

An illustration of the soft-mask module is shown in Figure~\ref{fig: soft-mask module}. The essential part of the soft-mask module is its dual-branch structure which contains a mask branch and an optical flow branch. The input feature maps represented as a set of volume feature vectors, $X \in \mathbb{R}^{s\times s \times c}$ are fed to both branches. The most significant contribution of this work is the separation of the optical flow field to multiple layers. For a separation into $k$ layers, $k$ masks will be generated in the mask branch as illustrated in Figure~\ref{fig: soft-mask module}. This requires $k$ convolutional filters $\{W_n^m, b_n^m\}_{n=1}^k$ in the mask branch. Correspondingly, the same number of filters are used in the optical flow branch $\{W_n^f, b_n^f\}_{n=1}^k$. The mask and intermediate optical flow are then computed as follows:

\begin{align}
\label{eqn: computation of masks and flows}
m_n =& X^T W_n^m + b_n^m &\! \text{for $n = 1\dots k$} \nonumber\\
f_n =& X^T W_n^f + b_n^f &\! \text{for $n = 1\dots k$}
\end{align}
Thus, given $k$ filters, we obtain $k$ corresponding pairs of mask and intermediate optical flow. By using $k$ filters in the optical flow branch and generating $k$ intermediate optical flow fields, we assume that each filter works independently and model a single type or a few types of object motions. Correspondingly, filters in the mask branch are expected to mask out parts with consistent motions by being high in certain regions and low in others. This leads us to use a maxout operation to extract mask entries with maximal activation along the channel axis. After the maxout operation, for each mask $m_n (n=1\dots k)$, all entries will be zero-out except for entries whose activation values are maximal in some regions among all masks. We denote the masks after maxout using $\{m_n'\}_{n=1}^k$. Following the maxout operation, there is no intersection among masks, and the union of all $m_n', n=1\dots k$ has activation in the full region. The maxout is given by:

\begin{equation}
\label{eqn: maxout}
m_n'=
\begin{cases}
m_n, & \text{if}\ m_n = \max\limits_{p=1\dots k}(m_p) \\
0, & \text{otherwise}
\end{cases}
\end{equation}
where $n = 1\dots k$. Note that the masks produced by maxout are not converted to binary values, thus resulting in soft-masks. By using soft masks, we can mask out irrelevant parts and prioritize values in each layer. In the proposed approach, masks generated by the maxout operation are applied to corresponding intermediate optical flow field by element-wise multiplication as shown below:

\begin{equation}
\label{eqn: fuse flow}
f_n'=
\begin{cases}
m_n' \times f_n, & \text{if}\ m_n' \neq 0 \\
0, & \text{otherwise}
\end{cases}
\end{equation}
where $n=1\dots k$ The result of the above computation is a set of disjoint optical flow layers, each of which represents a certain type of motion. An illustration of the soft-mask module works is shown in Figure~\ref{fig: softmask pipeline} and results of generated masks are shown in Figure~\ref{fig: maxout demo}.

\subsection{Quadratic Fitting of Optical Flow}
Objects moving in different ways, result in different types of motion. The underlying optical flows are non-linear and locally piecewise smooth. 

\begin{figure}[th]
\centerline{
\begin{tabular}{c}
  \resizebox{0.48\textwidth}{!}{\rotatebox{0}{
  \includegraphics{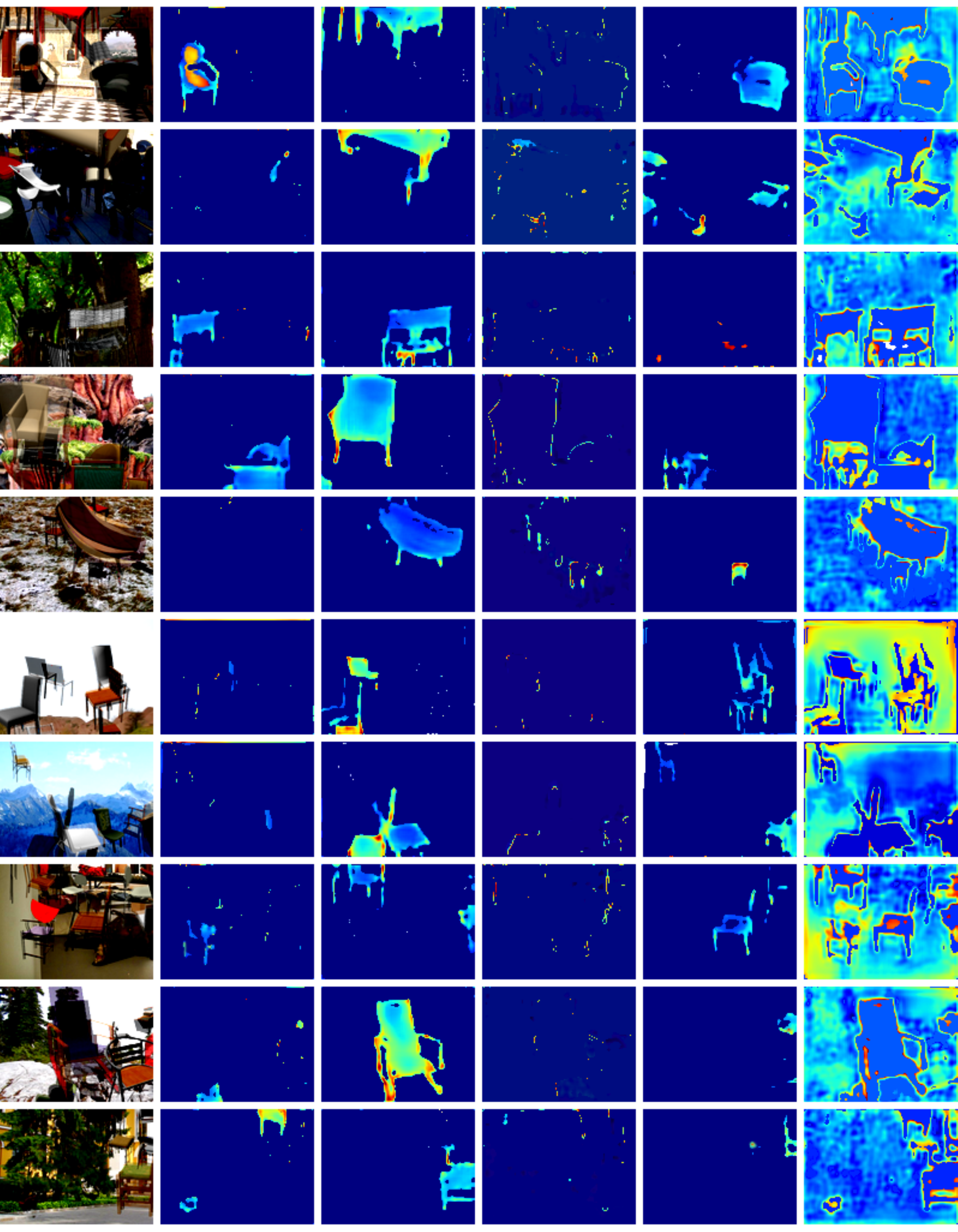}}}
  \\
\end{tabular}}
\caption{Examples of masks generated by the proposed soft-mask module. Five masks are generated for each input image pair which is shown as overlaid. Colors are according to normalized values in each image separately.}
\label{fig: maxout demo}
\end{figure}

There are two advantages in the proposed soft-mask module that make the estimation of optical flow easier. The first advantage is due to using maxout in the mask generation. By keeping only the maximal value among all masks, the optical flow is separated into multiple disjoint layers. The qualitative results as shown in Figure~\ref{fig: maxout demo} demonstrate that the soft-mask module allows the resulting masks to separate the flow field into pieces according to detected motion types. In addition, the masks detect the boundary of each motion piece, thus allowing the estimation of optical flow on boundaries to be more accurate.  The second advantage of using the proposed soft-mask module is that the output is quadratic in terms of feature maps $X$ fed to the module. To see this, consider the computation of masks and intermediate optical flow shown in Equation~\ref{eqn: computation of masks and flows}. The computation of non-zero $f_n'$ could be written as:

\begin{align}
f_n' & = m_n' \times f_n \nonumber\\
	 & = (X^T W_n^m + b_n^m) \times (X^T W_n^f + b_n^f) \nonumber\\
	 & = W_n^{mT} X X^T W_n^f + X^T(b_n^f W_n^m + b_n^m W_n^f) + b_n^m b_n^f
\end{align}
As can be observed in the above equation, the representation of $f_n'$ is quadratic in terms of the variable $X$.

\begin{figure}[h]
\centerline{
\begin{tabular}{ccc}
  \resizebox{0.14\textwidth}{!}{\rotatebox{0}{
  \includegraphics{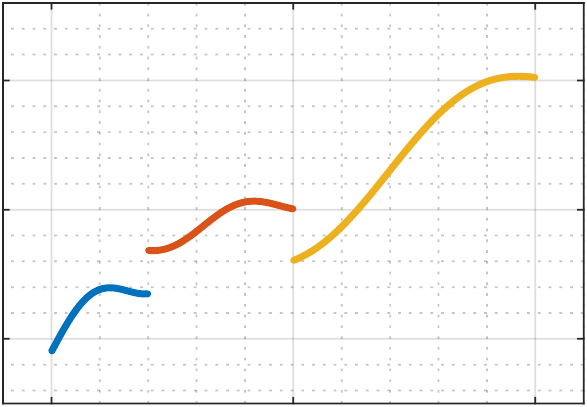}}}
  &
  \resizebox{0.14\textwidth}{!}{\rotatebox{0}{
  \includegraphics{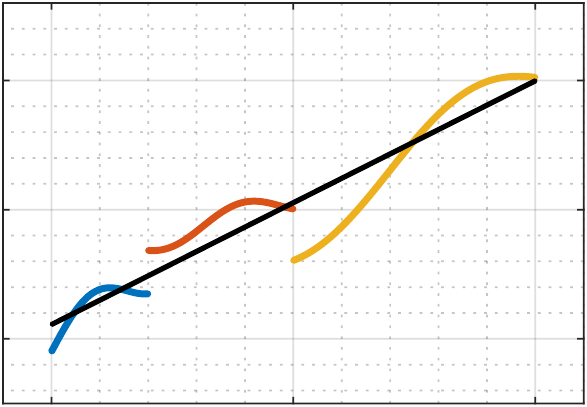}}}
  &
  \resizebox{0.14\textwidth}{!}{\rotatebox{0}{
  \includegraphics{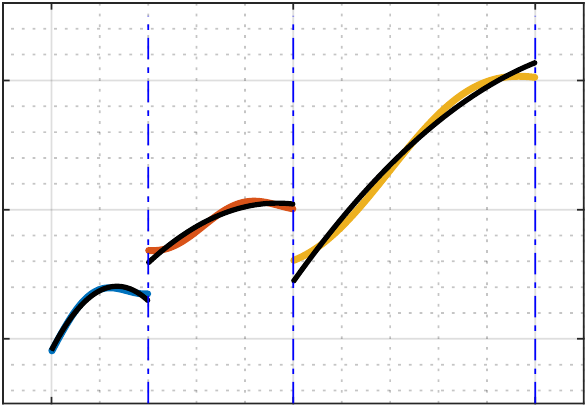}}}
  \\
  A & B & C
  \\
\end{tabular}}
\caption{A: Given data. B: Fitting using a linear function. C: Fitting using a piecewise quadratic function.}
\label{fig: quadratic demo}
\end{figure}

To better illustrate the advantage in using the soft-mask module with respect to linear output. Consider the 1D example shown in Figure~\ref{fig: quadratic demo}. In this example, function values are smooth in three separate domains. The improvement of fitting data using a piecewise quadratic function is shown in Figure~\ref{fig: quadratic demo} B and C.

\subsection{Regularization for Unsupervised Training}
Training an unsupervised neural network for optical flow estimation is possible by using a network similar to FlowNet for base optical flow inference followed by a spatial transform network (STN). To show that proposed soft-mask module can improve flow estimation using the same framework, we add the soft-mask module to FlowNet and use it as a base optical flow inference network in an unsupervised training framework.

The smoothness term which is used by all above unsupervised approaches plays a significant role in regularizing the local consistency of optical flow. We use the bending energy regularization~\cite{rohlfing2003volume}:

\begin{align*}
\varphi(\bold{u}, \bold{v}) = & \sum ((\frac{\partial^2\bold{u}}{\partial \bold{x}^2})^2+(\frac{\partial^2\bold{u}}{\partial \bold{y}^2})^2+2(\frac{\partial^2\bold{u}}{\partial \bold{x} \partial \bold{y}})^2) + \\ 
& \sum ((\frac{\partial^2\bold{v}}{\partial \bold{x}^2})^2+(\frac{\partial^2\bold{v}}{\partial \bold{y}^2})^2+2(\frac{\partial^2\bold{v}}{\partial \bold{x} \partial \bold{y}})^2)
\end{align*}
where $\bold{u}, \bold{v} \in \mathbb{R}^{H\times W}$ are estimated horizontal and vertical components of the optical flow field.

\section{Empirical Evaluation}
\label{sec: evaluation}
\subsection{Benchmark}
We evaluatethe our performance of the proposed approach on three standard optical flow benchmarks: Flying Chairs~\cite{7410673},  Sintel~\cite{Butler:ECCV:2012}, and KITTI~\cite{geiger2012we}. We compare the performance of the proposed approach to both supervised methods such as: FlowNet(S/C)~\cite{7410673}\cite{Ilg_2017_CVPR}, SPyNet~\cite{Ranjan_2017_CVPR}, as well as DeepFlow~\cite{weinzaepfel2013deepflow}, and EpicFlow~\cite{revaud2015epicflow}. We compare the proposed approach to methods  including: DSTFlow~\cite{ren2017unsupervised}, USCNN~\cite{ahmadi2016unsupervised}, and back-to-basic unsupervised FlowNet (bb-FlowNet)~\cite{DBLP:journals/corr/YuHD16}. 

Recently, FlowNet 2.0, a follow-up work of FlowNet, achieved state of the art results on most datasets. The architecture of FlowNet 2.0~\cite{Ilg_2017_CVPR} uses several FlowNets and contains cascade training of the FlowNets in different phases. Since the focus of this paper is on using the soft-mask module to boost performance of a single network, we do not include FlowNet 2.0 in our evaluation. Note that the proposed soft-mask module can be incorporated into FlowNet 2.0.

\subsection{Network Structure}
The goal of this paper is to show how the performance of existing optical flow networks can be improved by replacing the normal optical flow output layer with the proposed soft-mask module. We choose FlowNetS and FlowNetC as the base networks and replace their optical flow output layers with a soft-mask module. Using the layered optical flow estimation (LOFE) proposed in this paper, we term the resulting modified networks: FlowNetS+LOFE and FlowNetC+LOFE, respectively. To make our evaluation more complete, we also replaced the output layer of SPyNet with the soft-mask module. The resulting model is labeled as SPyNet+LOFE.

\begin{figure}[th]
\centerline{
\begin{tabular}{ccc}
  \resizebox{0.145\textwidth}{!}{\rotatebox{0}{
  \includegraphics{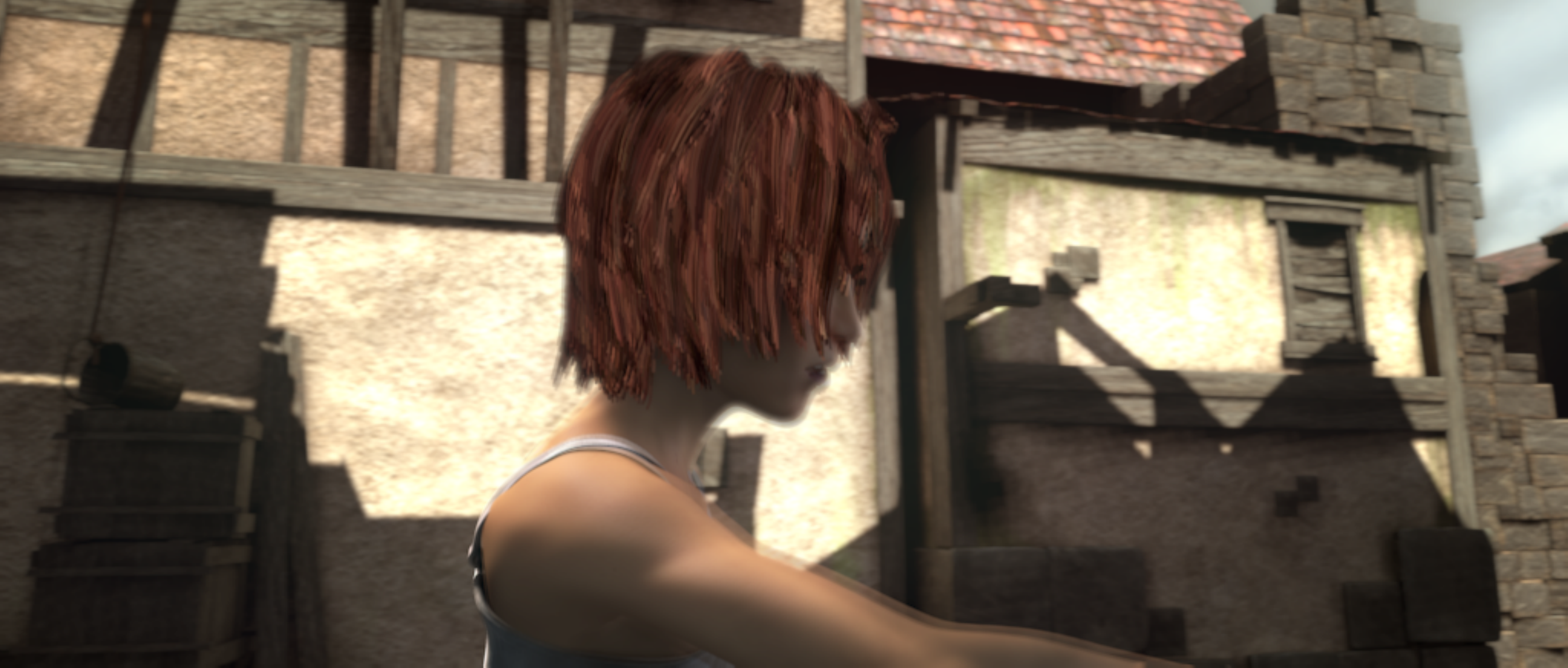}}}
  &
  \resizebox{0.145\textwidth}{!}{\rotatebox{0}{
  \includegraphics{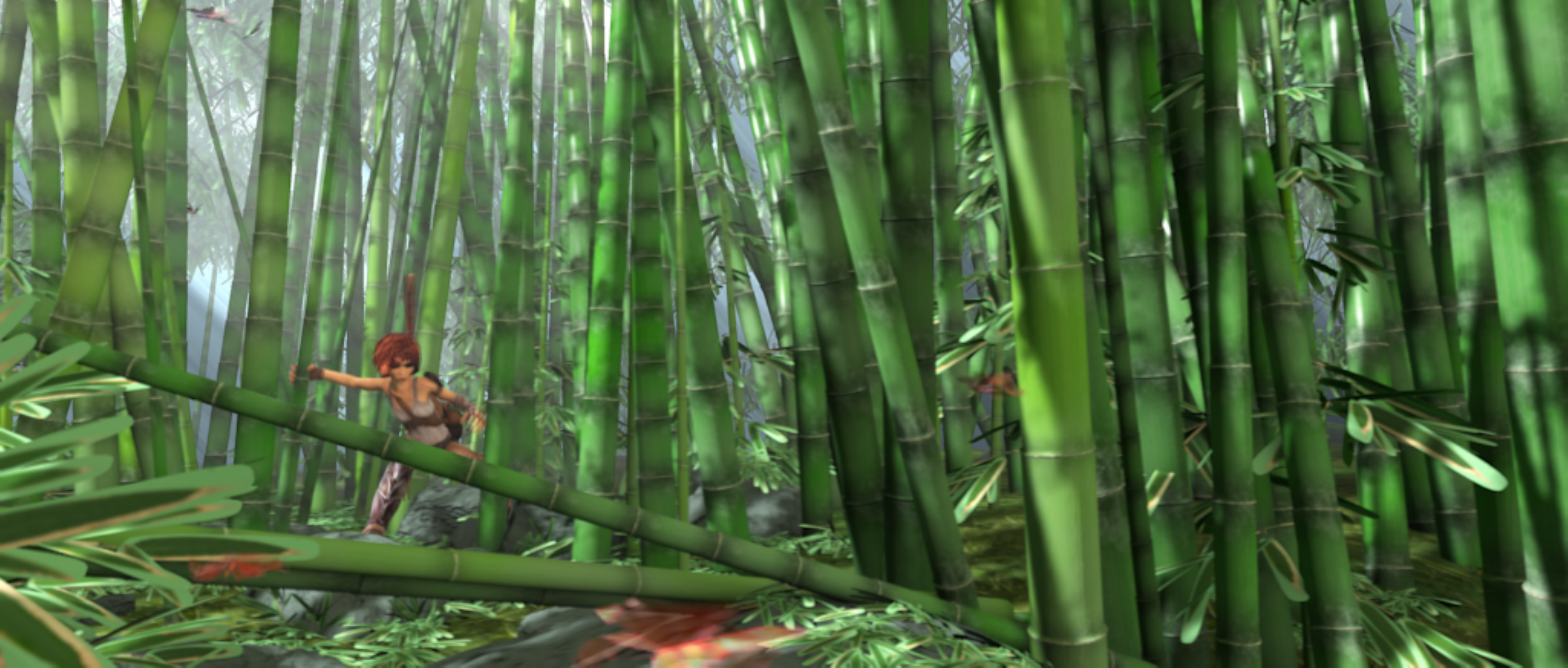}}}
  &
  \resizebox{0.145\textwidth}{!}{\rotatebox{0}{
  \includegraphics{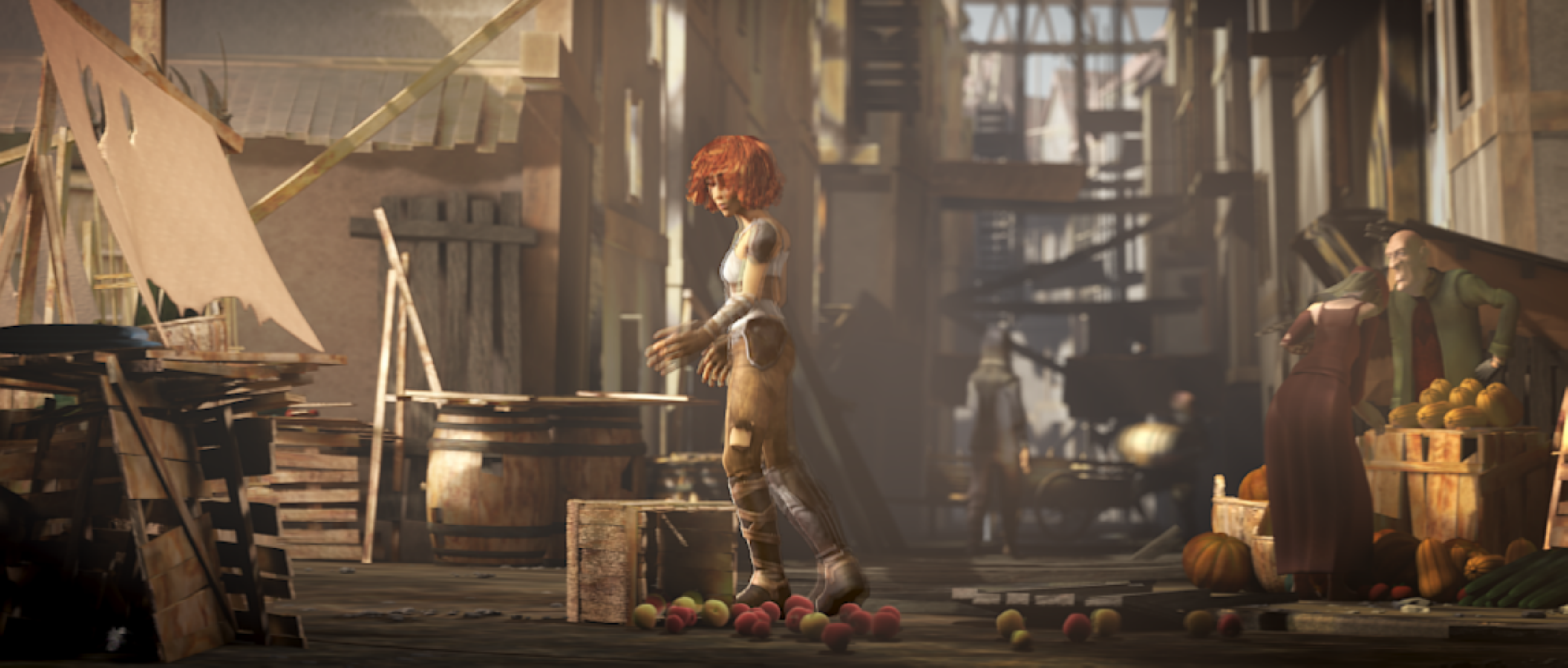}}}
  \\
  \resizebox{0.145\textwidth}{!}{\rotatebox{0}{
  \includegraphics{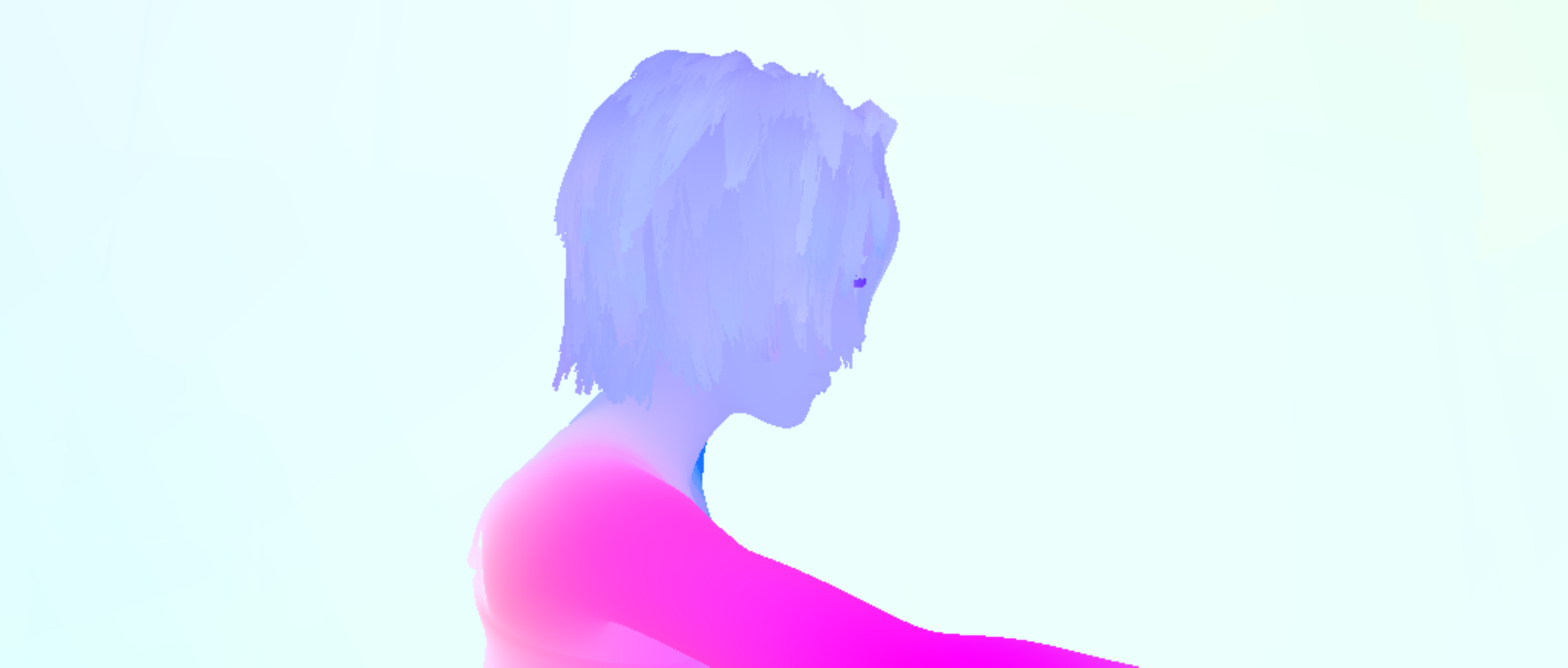}}}
  &
  \resizebox{0.145\textwidth}{!}{\rotatebox{0}{
  \includegraphics{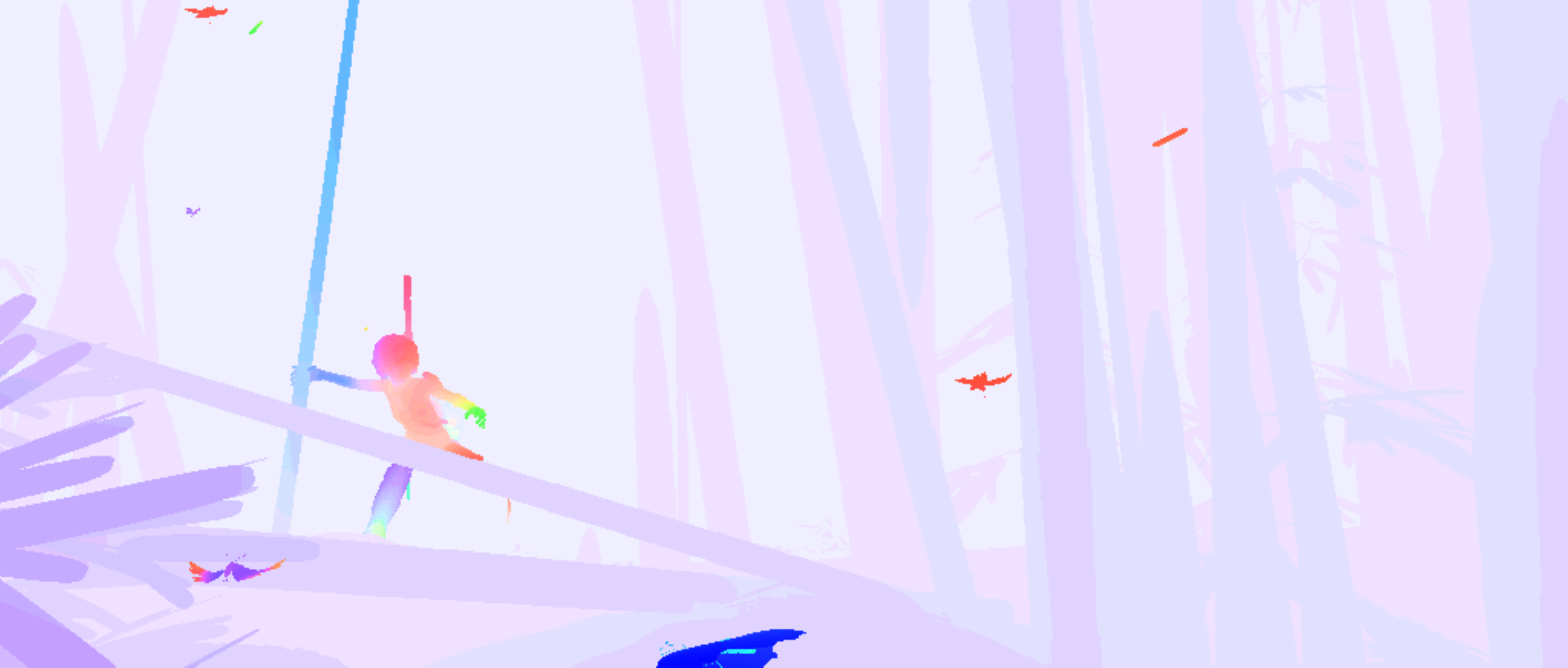}}}
  &
  \resizebox{0.145\textwidth}{!}{\rotatebox{0}{
  \includegraphics{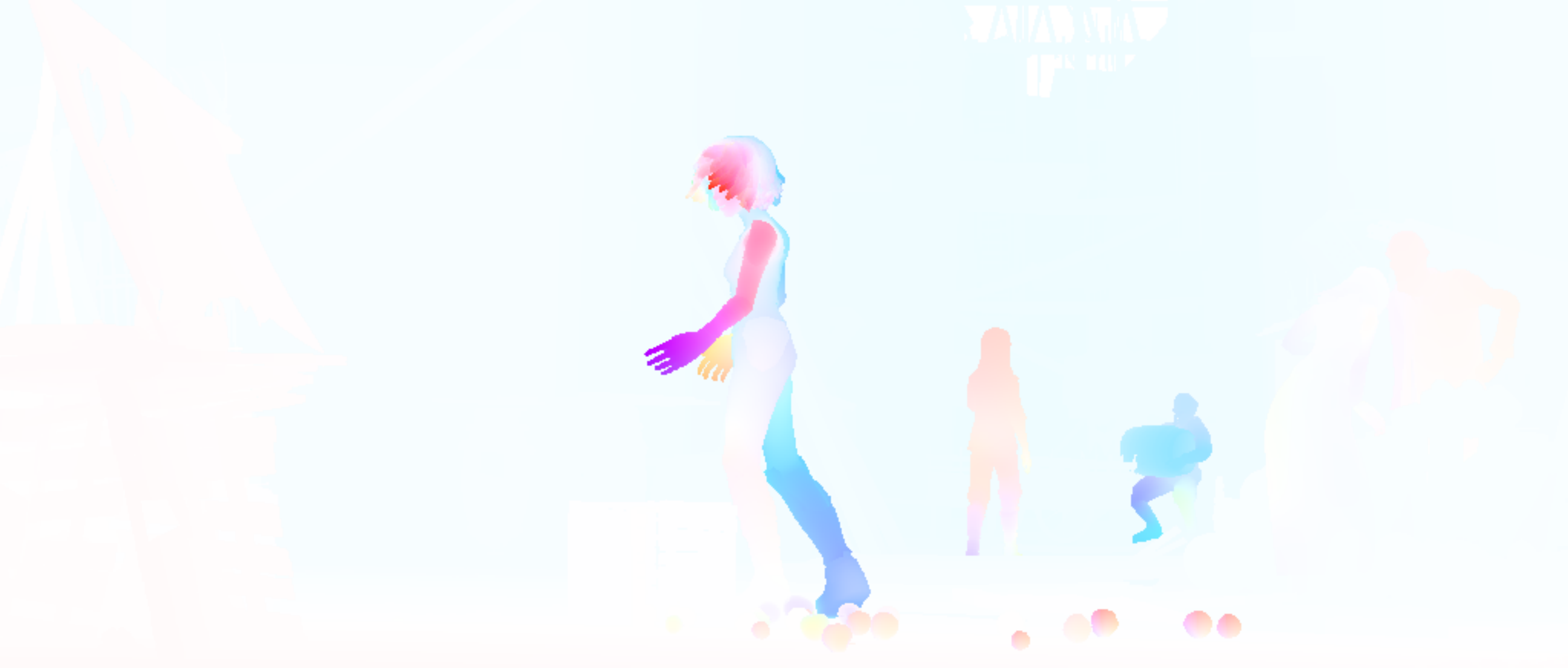}}}
  \\
  \resizebox{0.145\textwidth}{!}{\rotatebox{0}{
  \includegraphics{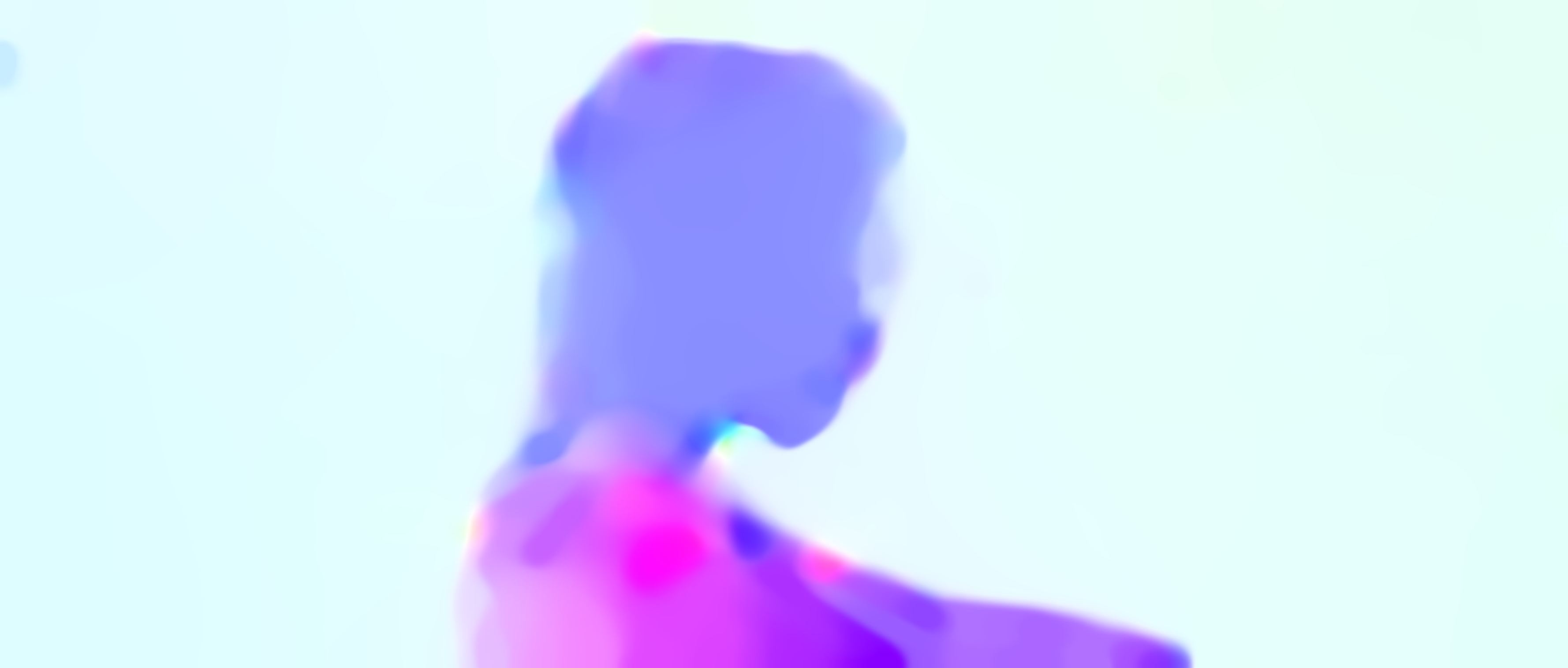}}}
  &
  \resizebox{0.145\textwidth}{!}{\rotatebox{0}{
  \includegraphics{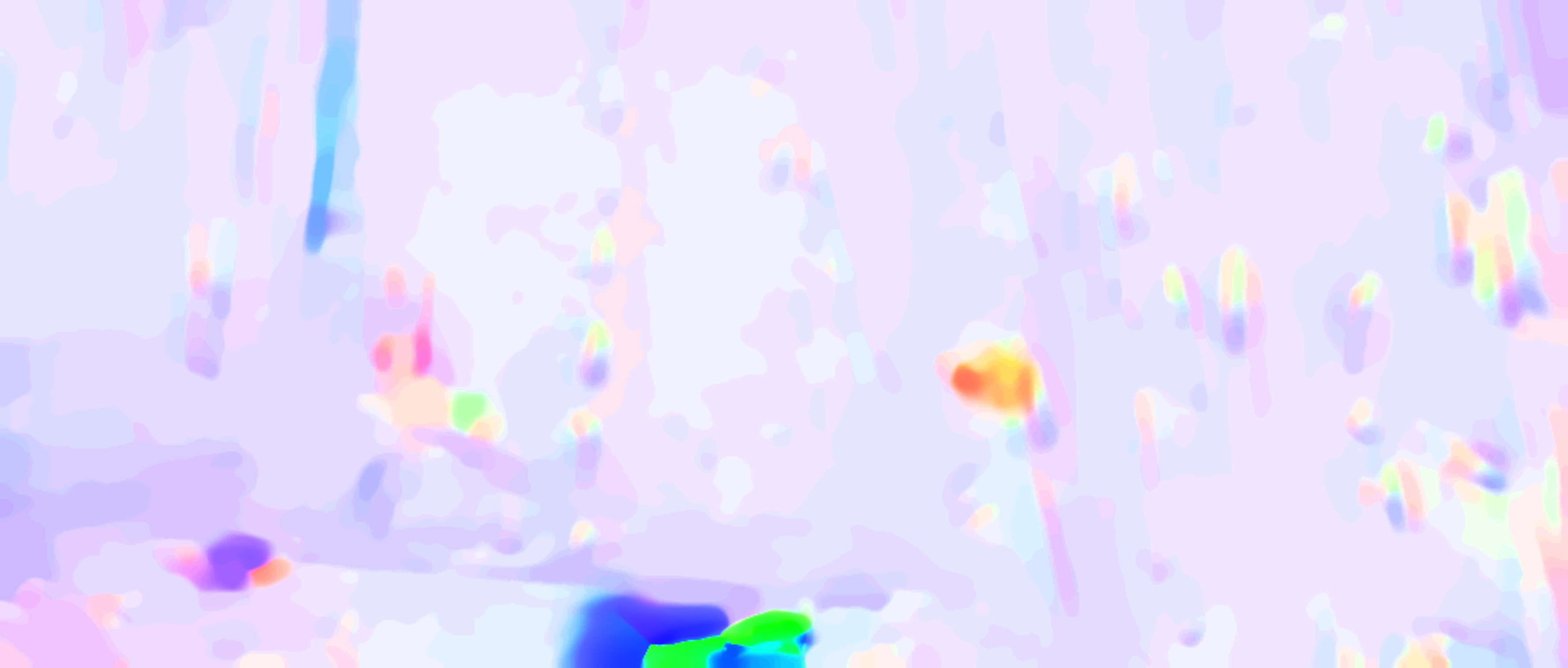}}}
  &
  \resizebox{0.145\textwidth}{!}{\rotatebox{0}{
  \includegraphics{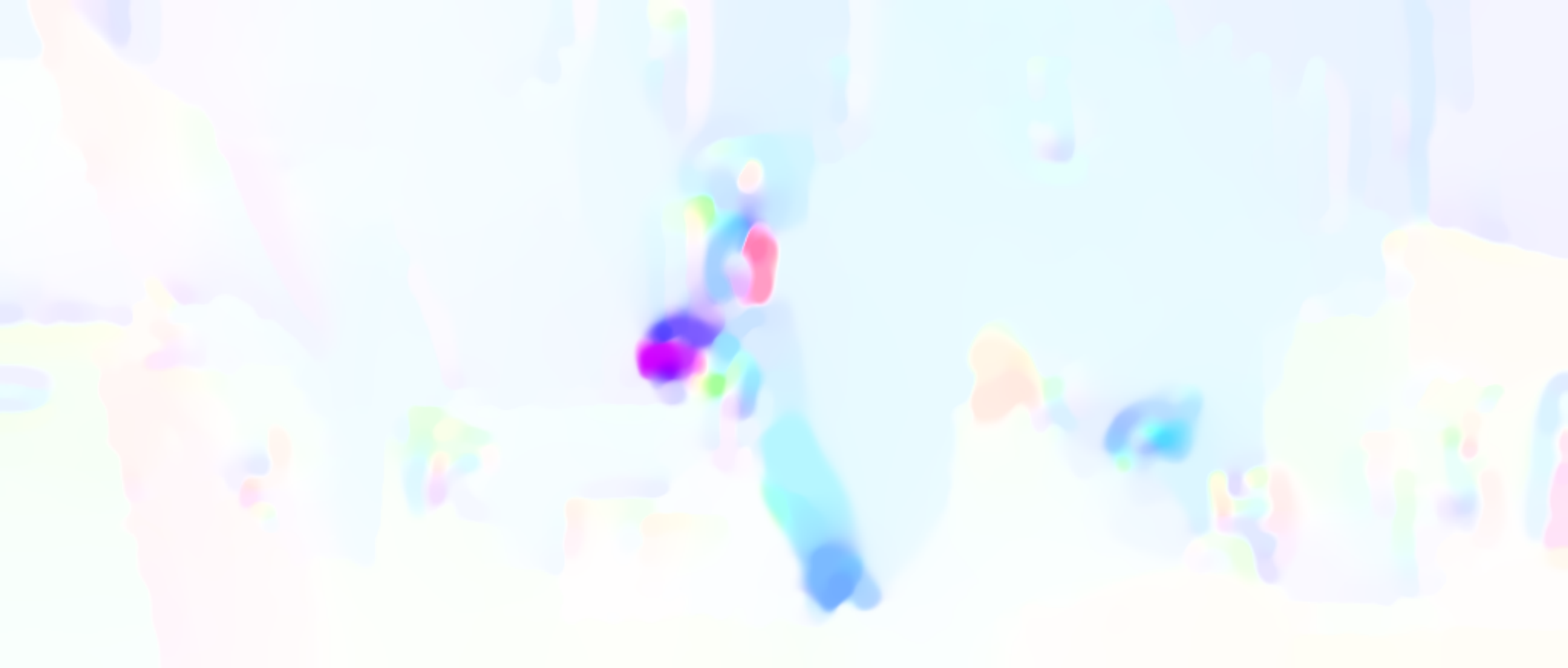}}}
  \\
  \resizebox{0.145\textwidth}{!}{\rotatebox{0}{
  \includegraphics{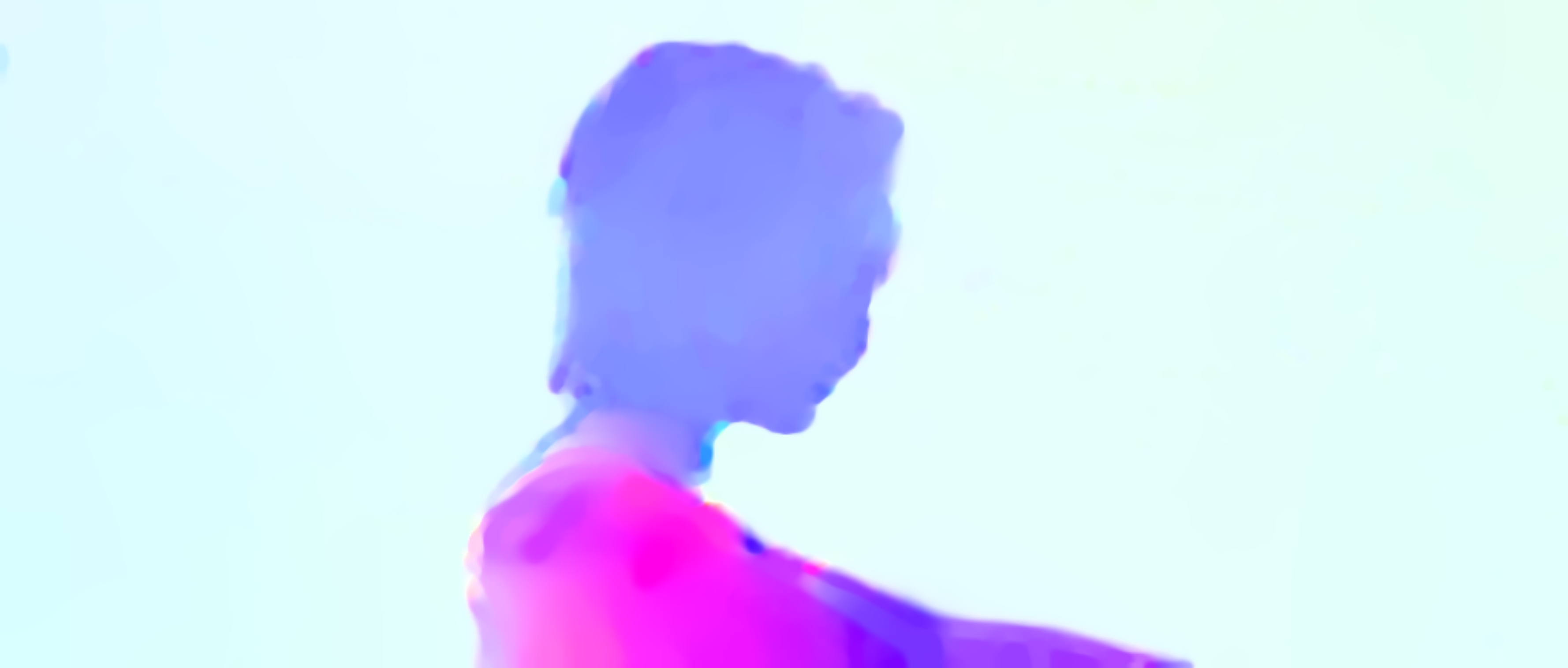}}}
  &
  \resizebox{0.145\textwidth}{!}{\rotatebox{0}{
  \includegraphics{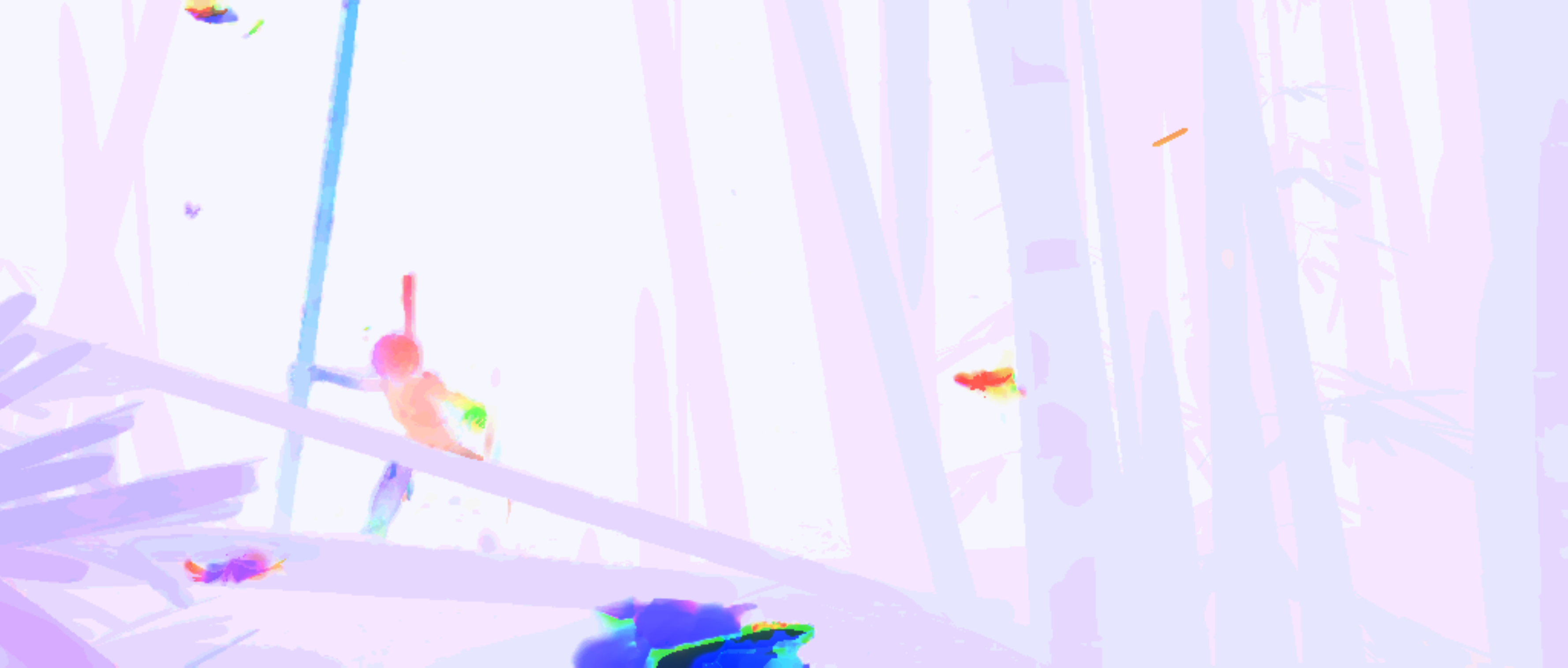}}}
  &
  \resizebox{0.145\textwidth}{!}{\rotatebox{0}{
  \includegraphics{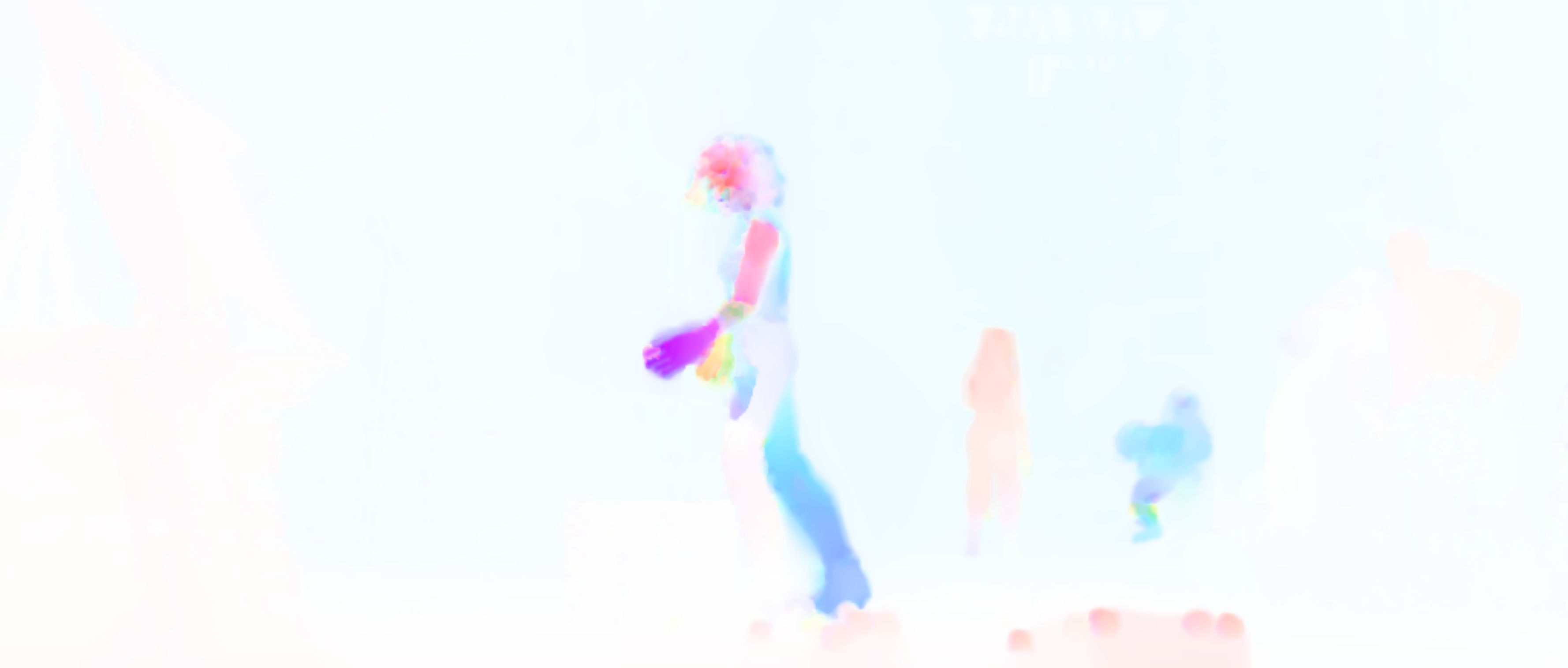}}}
\end{tabular}}
\caption{Examples of predicted flows compared with results from FlowNetC. First row: input image pair (overlaid). Second row: ground truth optical flow. Third row: flow results of FlowNetC. Fourth row: flow results of FlowNetC+LOFE.}
\label{fig: predicted flows}
\end{figure} 


\begin{table*}[th]
\centering
\caption{Average end point errors (EPE) of the proposed networks compared to several existing methods on Flying Chairs and Sintel Clean datasets. EpicFlow and DeepFlow are traditional methods which do not use neural networks. All other methods in the table are trained with supervised data. Bold font indicates the most accurate results among the network-based methods.}
\label{tab: results supervised}
\begin{tabular}{lccccccc}
\hline
\hline
\multicolumn{1}{c}{Method} & Flying Chairs & \multicolumn{2}{c}{Sintel Clean} & \multicolumn{2}{c}{Sintel Final} & \multicolumn{1}{c}{KITTI} & Time (s) \\
                           & Test          & Train           & Test           & Train           & Test           & Train 	&          \\ \hline
EpicFlow                   & 2.94          & 2.40            & 4.12           & 3.7             & 6.29           & 3.47    	& 16       \\
DeepFlow                   & 3.53          & 3.31            & 5.38           & 4.56            & 7.21           & 4.58    	& 17       \\ \hline
FlowNetS+schd              & 2.69          & 4.42            & 6.86           & 5.25            & 7.46           & 8.64    	& 0.12     \\
FlowNetC+schd              & 2.35          & 4.24            & 6.82           & 5.07            & 8.34           & 8.85    	& 0.23     \\
FlowNetS+schd+ft           & 2.49          & 4.08            & 6.98           & 4.75            & 7.52           & 8.26   	& 0.12     \\
FlowNetC+schd+ft           & 2.17          & 3.79            & 6.83           & 4.59            & 7.99           & 8.35     	& 0.23     \\
SPyNet                     & 2.63          & 4.23            & 6.82           & 5.67            & 8.49           & 9.12     	& 0.11     \\
SPyNet+LOFE				    & 2.33          & 3.99            & 6.52           & 5.30            & 8.49           & 9.12    	& 0.31     \\
FlowNetS+LOFE+schd        	& 2.49          & 4.20            & 6.80           & 4.88            & \textbf{7.36}           & 8.03   	& 0.36        \\
FlowNetC+LOFE+schd       	& 2.17          & 3.96            & 6.78           & \textbf{4.54}            & 7.59           & 8.14    	& 0.47     \\
FlowNetS+LOFE+schd+ft    	& 2.37          & 3.81            & 6.44           & 4.62            & 7.45           & \textbf{7.98}     	& 0.35        \\
FlowNetC+LOFE+schd+ft     	& \textbf{2.02} & \textbf{3.49}            & \textbf{6.21}           & 4.56            & \textbf{7.51}           & 8.01   	& 0.46     \\ \hline
\end{tabular}
\end{table*}

\subsection{Training Details}
\noindent\textbf{Training of Soft-mask Module.} Both FlowNetS+LOFE and FlowNetC+LOFE could be built by simply replacing the output layer with a soft-mask module. Data scheduling has been shown very useful when training FlowNetS and FlowNetC in \cite{Ilg_2017_CVPR}. For supervised networks, we used pre-trained weights of two networks from FlowNet 2.0 \cite{Ilg_2017_CVPR} both trained with and without fine-tuning. Then, for each dataset in our experiment, we trained the soft-mask module by fixing the pre-trained weights. We compared our proposed method to FlowNetS and FlowNetC trained using data scheduling. We are aware that it is unfair by training soft-mask module more. Therefore, when compared to baseline models, we trained their output layers and fixed other parts in the same way.
\newline
\newline
\noindent\textbf{Data augment.} Various types of data augmentation are used during training. We applied rotation at random within $[\ang{-17}, \ang{17}]$. A random translation within $[-50, 50]$ pixels was applied to both horizontal and vertical directions. In addition, following~\cite{Ranjan_2017_CVPR} we included additive white Gaussian noise sampled uniformly from $\mathcal{N}(0, 0.1)$. We also applied color jitter with additive brightness, contrast and saturation sampled from a Gaussian, $\mathcal{N}(0, 0.4)$. All data augmentations were done using GPU during training.

\subsection{Results}
Evaluation was done with compared methods in two groups according to whether the training of the method is unsupervised or supervised. Table~\ref{tab: results supervised} shows the endpoint error (EPE) of the proposed network and several well-known methods. The endpoint error measures the distance in pixels between known optical flow vectors and estimated ones. Except for EpicFlow and DeepFlow, all other methods were trained in a supervised manner. We compare results of unsupervised methods in Table~\ref{tab: results unsupervised}.
\newline
\newline
\noindent \textbf{Supervised methods.} The proposed FlowNetS+LOFE and FlowNetC+LOFE tested in this group use $k=10$ for the number of layers. As can be seen in Table~\ref{tab: results supervised}, FlowNetS+LOFE and FlowNetC+LOFE achieve better performance compared with methods not using the module. We observe that the performance of SPyNet is also boosted by replacing the optical flow output layer with the soft-mask module. Considering the computation time we observe a small time increment when using the soft-mask module, and which is in an acceptable range. 

\begin{table*}[th]
\centering
\caption{EPE errors of methods that are trained without supervision. The results of compared methods are taken directly from the corresponding paper. The notation `ft' means fine-tuning. }
\label{tab: results unsupervised}
\begin{tabular}{lccccccc}
\hline
\hline
\multicolumn{1}{c}{Method} & Flying Chairs & \multicolumn{2}{c}{Sintel Clean} & \multicolumn{2}{c}{Sintel Final} & \multicolumn{2}{c}{KITTI}      \\
                           &               & Train           & Test           & Train          & Test            & Train          & Test          \\ \hline
DSTFlow                    & 5.11          & 6.93            & 10.40          & 7.82           & 11.11           & \textbf{10.43} & -             \\
USCNN                      & -             & -               & -              & 8.88           & -               & -              & -             \\
BB-FlowNet                 & 5.36          & -               & -              & -              & -               & 11.32          & \textbf{9.93} \\
FlowNetS+LOFE              & \textbf{4.81} & \textbf{6.56}   & 10.10          & \textbf{7.62}  & 10.98           & 10.78          & 10.82         \\
FlowNetC+LOFE              & 4.92          & 6.78            & \textbf{9.98}  & 7.77           & \textbf{10.19}  & 11.01          & 11.25         \\ \hline
\end{tabular}
\end{table*}

Qualitative results are shown in Figure~\ref{fig: predicted flows}. It could be observed that due to the soft-mask module in FlowNetC+LOFE,  generally the network has a better prediction on flow boundaries over FlowNetC.
\newline
\newline
\noindent \textbf{Unsupervised methods.} Training optical flow estimation networks without supervision is straight forward. The results are shown in Table~\ref{tab: results unsupervised}. As can be observed, the proposed networks achieve the best performance except for the KITTI dataset where the proposed approach achieved 2nd place. 

\subsection{Evaluation of the Soft-mask Module}
\label{sec: evaluation of the soft-mask module}
Since we replace the simple linear output layer in FlowNet(S/C) with a more complex soft-mask module, we would like to verify whether the improved results are obtained due to the way the soft-mask module works and not simply due to having a model with more coefficients. To better investigate the operation of the soft-mask module, we compared the FlowNetC+LOFE with three other networks in which we slightly changed the structures of the soft-mask module.

\begin{table}[h]
\centering
\caption{Comparison of the proposed FlowNetC+LOFE and its three variants. The notation ‘schd’ represents that networks are trained using data scheduling}
\label{tab: flownet variants}
\begin{tabular}{lcc}
\hline
\hline
                         & \thead{Chairs}  	& \thead{Sintel} \\ \hline
FNetC+schd                 & 2.35 				& 4.24             \\
FNetC+LOFE/no-maxout+schd  & 2.29  				& 4.12             \\
FNetC+LOFE/normalize+schd  & 2.32  				& 4.03             \\
FNetC+LOFE/no-masks+schd   & 2.62  				& 4.35             \\
FNetC+LOFE+schd            & \textbf{2.17}  				& \textbf{3.96}             \\
\hline
\end{tabular}
\end{table}

In the first network, given the proposed structure as FlowNetC+LOFE, we removed the maxout operation from the soft-mask module and kept the remaining configuration the same. We denote the resulting work FlowNetC+LOFE/no-maxout. In this case, FlowNetC+LOFE/no-maxout will have the exact same number of coefficients as FlowNetC+LOFE.  In \cite{zhou2016view}\cite{flynn2016deepstereo}, intermediate generated masks are also combined with extracted image features. Instead of adopting a max-out operation, masks are normalized in their works. Therefore, for the second network, we first copied the structure of FlowNetC+LOFE/no-maxout and employed the same normalization in the second network. We denote the resulting network as FlowNetC+LOFE/normalize. For the third network, we removed the mask branch from the soft-mask module and left the intermediate optical flow only. The third network is denoted as FlowNetC+LOFE/no-masks. 

For all four networks, we used $k=10$ in the soft-mask module. We used FlowNetC trained by data scheduling without fine-tuning as a baseline in the evaluation. To obtain an unbiased evaluation result, we trained and tested each of these networks on both Flying Chairs and Sintel dataset~\cite{Butler:ECCV:2012} three times. The average EPE is reported in Table~\ref{tab: flownet variants}.

As we can see from Table~\ref{tab: flownet variants}, the proposed FlowNetC+LOFE performed better than its three variants. This comparison leads to three conclusions. First, the better performance obtained by adding the soft-mask module to FlowNetC is not because using a larger model. Since both no-maxout and normalize versions of the proposed network have the identical complexity to the proposed network. Thus we conclude that the maxout operation makes optical flow estimation a more manageable task by separating optical flows into multiple layers. Second, with the same structure, FlowNetC+LOFE is better than no-maxout and normalize versions of the model. This result is caused by the maxout operation in the proposed network which can separate flows to layers to better generate flows in local regions. Third, the performance of FlowNetC/no-maxout and FlowNetC/normalize are both better than FlowNetC/no-masks version. While a possible hypothesis is that the model of no-maxout is larger than the model of no-masks. However, since FlowNetC, the smallest model in this comparison, achieved a better performance compared with the no-masks model. We thus conclude that the actual reason is a fact that both the FlowNetC/no-maxout and the FlowNetC/normalize models are using a quadratic function to fit optical flow instead of the linear function which is used in the no-masks FlowNetC model.


\begin{table}[]
\centering
\caption{EPE as a function of k, the number of masks generated intermediately.}
\begin{tabular}{llllll}
\hline
\hline
k value      & 5     & 10     & 20      & 30      & 40    \\ \hline
EPE      & 2.192 & 2.173 & 2.176 & 2.237 & 2.249 \\ \hline
\end{tabular}
\label{tab: evaluation k}
\end{table}

We investigate the relationship between $k$ the number of masks and flow layers used in the soft-mask module, and network performance in terms of EPE. Experiments were done using the Flying Chairs dataset. We set $k=5\mathrm{x}$, where $\mathrm{x}=1, \dots, 8$. As can be observed in Table~\ref{tab: evaluation k}, there is an immediate benefit to using the soft-mask module with respect to FlowNetC, where $k=5$ will efficiently boost performance. We see a convergence of EPE after $k=10$ and a slightly increase when $k>20$. This may be due to slight overfitting when separating the optical flow to too many layers.

\section{Conclusion}
We describe a new approach for optical flow estimation by combining a traditional layered flow representation with a deep learning method. Rather than pre-segmenting images to layers, the proposed approach automatically learns a layered representation of optical flow using the proposed soft-mask module. The soft-mask module has the advantage of splitting flow to layers in which the computation of the flow is quadratic in terms of input features. For evaluation, we use FlowNet as our base net to add the soft-mask module. The resulting networks are tested on three well-known benchmarks with both supervised and unsupervised flow estimation tasks. Experimental results show that the proposed network achieves better results with respect to the original FlowNet. 

\bibliographystyle{named}
\bibliography{mybib}

\end{document}